\documentclass{bmvc2k}

\usepackage[utf8]{inputenc} 
\usepackage[T1]{fontenc}    
\usepackage{hyperref}       
\usepackage{url}            
\usepackage{booktabs}       
\usepackage{amsfonts}       
\usepackage{nicefrac}       
\usepackage{microtype}      
\usepackage{xcolor}         
\usepackage{float}
\usepackage{graphicx}
\usepackage{dsfont}
\usepackage{amsmath,amssymb}
\usepackage{xspace}
\usepackage{wrapfig}
\usepackage{multirow}
\usepackage{pifont}
\usepackage{lipsum}
\usepackage{rotating}
\usepackage{algorithm}
\usepackage{algpseudocode}

\title{Compressing Video Calls using Synthetic Talking Heads}

\addauthor{Madhav Agarwal}{madhav.agarwal@research.iiit.ac.in}{1}
\addauthor{Anchit Gupta}{anchit.gupta@research.iiit.ac.in}{1}
\addauthor{Rudrabha Mukhopadhyay}{radrabha.m@research.iiit.ac.in}{1}
\addauthor{Vinay P. Namboodiri}{vpn22@bath.ac.uk}{2}
\addauthor{C V Jawahar}{jawahar@iiit.ac.in}{1}

\addinstitution{
 CVIT, IIIT-Hyderabad, India
}
\addinstitution{
 University of Bath, England
}

\runninghead{Agarwal et al.}{Compressing Video Calls using Synthetic Talking Heads}

\begin{document}

\maketitle

\begin{abstract}
  

We leverage the modern advancements in talking head generation to propose an end-to-end system for talking head video compression. Our algorithm transmits pivot frames intermittently while the rest of the talking head video is generated by animating them. We use a state-of-the-art face reenactment network to detect key points in the non-pivot frames and transmit them to the receiver. A dense flow is then calculated to warp a pivot frame to reconstruct the non-pivot ones. Transmitting key points instead of full frames leads to significant compression. We propose a novel algorithm to adaptively select the best-suited pivot frames at regular intervals to provide a smooth experience. We also propose a frame-interpolater at the receiver's end to improve the compression levels further. Finally, a face enhancement network improves reconstruction quality, significantly improving several aspects like the sharpness of the generations. We evaluate our method both qualitatively and quantitatively on benchmark datasets and compare it with multiple compression techniques. We release a demo video and additional information at \url{https://cvit.iiit.ac.in/research/projects/cvit-projects/talking-video-compression}. 
\end{abstract}

\section{Introduction}

As we progress through the 21st century, the world continues to grow digitally and becomes more connected than ever! Video calls are a big part of this push and are a staple form of communication. The pandemic in 2020 led to a massive reduction in social interaction and fast-tracked its adoption. Universities and schools were forced to use video calls as the primary means of teaching, while for many, video calling remained the only way to connect with friends and family. While the number of video calls will continue to rise in the future, increasing bandwidth is a daunting task. Incidentally, over half the world's countries do not even have 4G services~\footnote{\url{https://en.wikipedia.org/wiki/List_of_countries_by_4G_LTE_penetration}}! Therefore, introducing video compression schemes to reduce the bandwidth requirement is a need of the hour. 

\paragraph{Traditional Video Compression Techniques}
Compressing video information has fascinated researchers for nearly a century. The first works dealt with analog video compression and were released in 1929~\cite{first}. A significant breakthrough in modern video compression was achieved by~\cite{dct} using a DCT-based compression technique leading to the first practical applications. This was followed by the widely adopted H.264~\cite{h_264} and H.265~\cite{h_265} video codecs, which remain the most popular in industrial applications. The most recent codec to be released is H.266~\cite{h_266}. However, we do not compare our work with H.266 due to the lack of availability of open-source implementations. Deep learning-based video compression techniques like~\cite{dlc1, dlc2, dlc3, dlc4} have also been prevalent in the recent past. These techniques use autoencoder-like structures to encode video frames in a bottlenecked latent space and generate it back on the receiver's end. While such approaches have proven their effectiveness in multiple situations, they are generic and do not consider the high-level semantics of the video for compression.

\paragraph{Talking Head Video Compression}
Video calls, on the other hand, encompass a specific class of videos. They primarily contain videos of speakers and are popularly known as talking head videos. The inherent semantic information present in a talking head video involving the face structure, head movements, expressions on the face, etc., has long interested researchers in developing compression schemes targeted towards such specialized videos. Techniques like~\cite{face_vid} transmit $68$ facial landmarks for each frame, which synthesize the talking head at the receiver's end. In 2021, Wang et al.~\cite{wang2021one} proposed using face reenactment for video compression. They used $10$ learned 3D key points instead of pre-defined face landmarks to represent a face in their work leading to significant compression. Each learned key point contains information regarding the structure of the face, rotation, translation, etc., and helps to warp a reference frame. 

\begin{figure*}[h]
    \centering
    \includegraphics[width=0.9\textwidth]{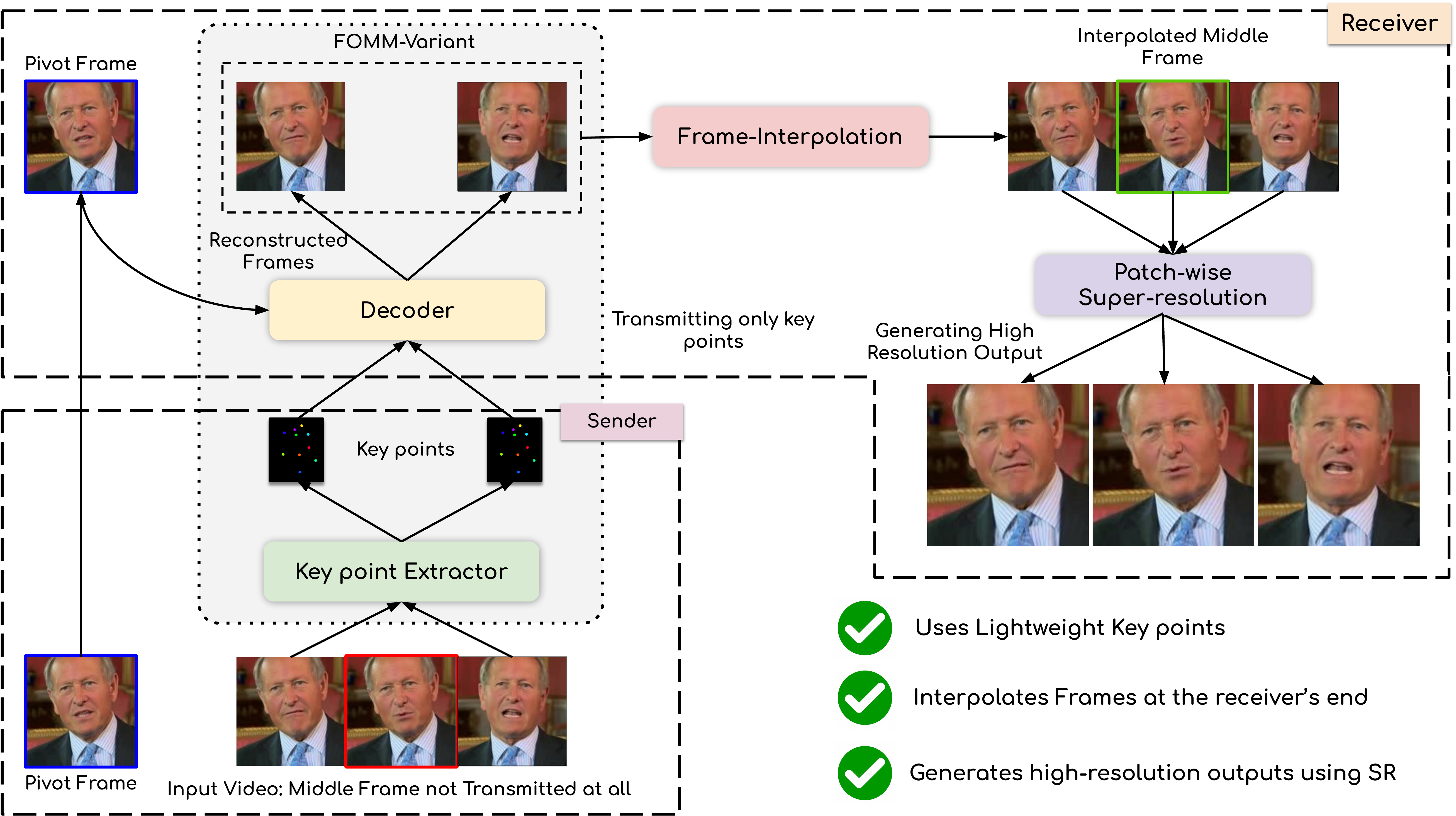}
    \caption{We depict the entire pipeline used for compressing talking head videos. In our pipeline, we detect and send key points of alternate frames over the network and regenerate the talking heads at the receiver's end. We then use frame interpolation to generate the rest of the frames and use super-resolution to generate high-resolution outputs.}
    \label{fig:pipeline}
\end{figure*} 


\paragraph{Our Contributions}
We explore this concept further in this work and propose several novel improvements. We first send a high-resolution frame (pivot frame) at the start of the video calls. For the rest of the frames, we use a modified version of~\cite{Siarohin_2019_NeurIPS} to detect key points in each of them and transmit them to the receiver. The key points are then used to calculate a dense flow that warps the pivot frame to recreate the original video. While~\cite{Siarohin_2019_NeurIPS, wang2021one} used $24$ bytes to represent a single key point, we further propose to reduce this requirement to only $8$ bytes. Next, we use a novel talking head frame-interpolater network to generate frames at the receiver's side. This allows us to send key points from fewer frames while rendering the rest of the frames using the interpolater network. We use a patch-wise super-resolution network to upsample the final outputs to arbitrary resolutions, significantly improving the generations' quality. In a lengthy video call sending a single pivot frame at the start of the video may lead to inferior results on significant changes in the background and head pose. Therefore, we also propose an algorithm to adaptively select and send pivot frames negating the effects of such changes. Overall, our approach allows for unprecedently low Bits-per-Pixel (BPP) value (bits used to represent a pixel in a video) while maintaining usable quality. We refer the reader to check our project web-page for numerous example results from our approach.

\section{Background: Synthetic Talking Head Generation}
\label{sec:format}

Our work revolves around synthetic talking head generation. Therefore, we survey the different types of talking head generation works prevalent in the community. Talking head generation was first popularized in works like~\cite{suwajanakorn2017synthesizing, you-said-that, jamaludin2019you-said-that-journal, lipgan, wav2lip} which attempted to generate only the lip movements from a given speech. These works were effective for solutions that required preserving the original head movements in a talking head video while changing only the lip synchronization to a new speech. A separate class of works~\cite{zhou2020makelttalk, zhang2021flow, wang2021audio2head, zhou2021pose} tried to generate the talking head video directly from speech without additional information. While these works can also potentially find their usage in video call compression, the head movements in the generated video do not match those of the original one, limiting its usage!

\paragraph{Face Reenactment}
In face reenactment, a source image is animated using the motion from a driving video. The initial models for this class of works were 
speaker-specific~\cite {bansal2018recycle,wu2018reenactgan}. These models are specifically trained on a single identity and cannot generalize to different individuals. On the other hand, speaker agnostic models~\cite{Siarohin_2019_NeurIPS,wang2021one,zhou2021pose, avfr2023wacv} are more robust. They require a single image of any identity and a driving video (need not have the same identity) to generate a talking head of the source identity following the driving motion. We find face reenactment works to be well suited for talking head video compression. We propose to use the inherent characteristic of the problem and send a single high-quality frame that can be animated by the rest of the video at the receiver's end to generate the final output. The reenactment is driven by landmarks, feature warping, or latent embeddings. First-Order-Motion-Model (FOMM) proposed by Siarohin et al.~\cite{Siarohin_2019_NeurIPS} uses self-learned key points to represent the dense motion flow of driving video. Each key point consists of the coordinates and Jacobians representing the local motion field between the source image and the driving video. A global motion field is then interpolated from the local motion field, and the source image is warped using the estimated motion field. Wang et al.~\cite{wang2021one} too similarly learn a motion field between the source image and the driving video. However, in their case, the key points are 3-dimensional, containing additional rotation and translation information.


\section{Methodology}
\label{sec:majhead}

\paragraph{Overview of the Technique}
As discussed previously, we start the video call by sending a pivot frame from the sender to the receiver and then animate it using the rest of the frames in the video call. We use a variation of the FOMM~\cite{Siarohin_2019_NeurIPS} model for achieving this task. Each key point in the FOMM model consists of 2D coordinates and Jacobians that possess additional region-specific information. Through experiments, we realize that Jacobians play an essential role in modeling complex motions. However, video calls are frontal face videos with relatively fewer head motions. Thus, we reduce the bits required to store each key point by removing Jacobians and transferring only the coordinates of key points for each frame over the network. We also propose a talking face frame interpolation algorithm inspired by~\cite{gpen} to generate intermediate frames in the video, reducing the number of frames for which key points needs to be transferred. Finally, we use a patch-based super-resolution network to generate arbitrary high-resolution outputs. To counter the instability caused by the removal of Jacobians when encountered with large head movements, we formulate a simple algorithm to send and replace pivot frames intermittently based on the difference in head pose and background between the current pivot frame and the driving video at the sender's side.


\paragraph{Formalizing the Compression Strategy}
Let us assume we have $n+1$ frames at the sender's end in our setup. We denote the frames by $f_0, f_1, f_2, ..., f_n$. We pick $f_0$ as our first pivot frame and transmit it to the receiver. The pivot frame is denoted by $f_{pv}$. We then pick alternate frames $f_1, f_3, f_5,  ...$ and pass them through the learned key point detector of our FOMM-variant. The detected key points are denoted by $p_1, p_3, p_5, ...$. At the receiver's end, the decoder from the FOMM-variant uses the transmitted key points and $f_{pv}$ to generate $f^{\prime}_{1}, f^{\prime}_{3}, f^{\prime}_{5}, ...$. We use our frame-interpolater network to generate the intermediate frames, $f^{\prime}_{2}, f^{\prime}_{4}, ...$. We then apply our patch-based super-resolution network on all the frames on the receiver's end, $f^{\prime}_{1}, f^{\prime}_{2}, f^{\prime}_{3}, f^{\prime}_{4}, f^{\prime}_{5},...$ to generate higher resolution versions of the same. Finally, for a significant difference in the head pose or background between the pivot frame $f_{pv}$ and the $i^{th}$ frame, $f_i$, we transmit $f_i$ to the receiver making it the new pivot frame.   


\paragraph{Modifying the First-Order-Motion-Model}
We take inspiration from First Order Motion Model for Image Animation~\cite{Siarohin_2019_NeurIPS} for reenacting a face at the receiver's end. 
While the original version of FOMM~\cite{Siarohin_2019_NeurIPS} was not designed for compression in video calls, we re-purpose it for the task at hand and built a refined version of the model. 
In the original model, a key point detector detects $10$ key points along with Jacobians in the neighborhood of each key point. The model detects these key points in both the source and driving frames and a motion field is calculated between corresponding key points between the two frames. The dense flow calculated from this motion field is then used to warp the source frame using a decoder generating the final output. All the network components like the generator and the key-point detector are trained end-to-end allowing the key-point detector to extract key points best suited for generating the most accurate result. 

In this work, we remove the requirement of Jacobians and instead train a version of FOMM\footnote{\url{https://github.com/AliaksandrSiarohin/first-order-model}} requiring only coordinates of the key points to reconstruct a frame. This is motivated directly by our use-case of video call compression. Jacobians are $2\times2$ integer matrices for each of the $10$ key points. By removing the Jacobians, we can represent a frame with only the $(x, y)$ coordinates of the $10$ key points saving a large amount of bandwidth. We find that removing the Jacobians does not affect the performance of our network on frontal-facing videos that are most encountered during a video call. We follow the same training methodology and losses as stated in~\cite{Siarohin_2019_NeurIPS} to train this modified version of the FOMM model. Once the model is trained, the key point extractor is deployed at the sender's end while the decoder part of the network is deployed at the receiver's end. At any point of the video call, the current pivot frame acts as the source frame, and the key points from the subsequent frames (which serve as the driving video) are used to warp the pivot frame animating it. A graphical representation of the process is given in Figure~\ref{fig:pipeline}.

\begin{figure*}[h]
    \centering
    \includegraphics[width=0.9\textwidth]{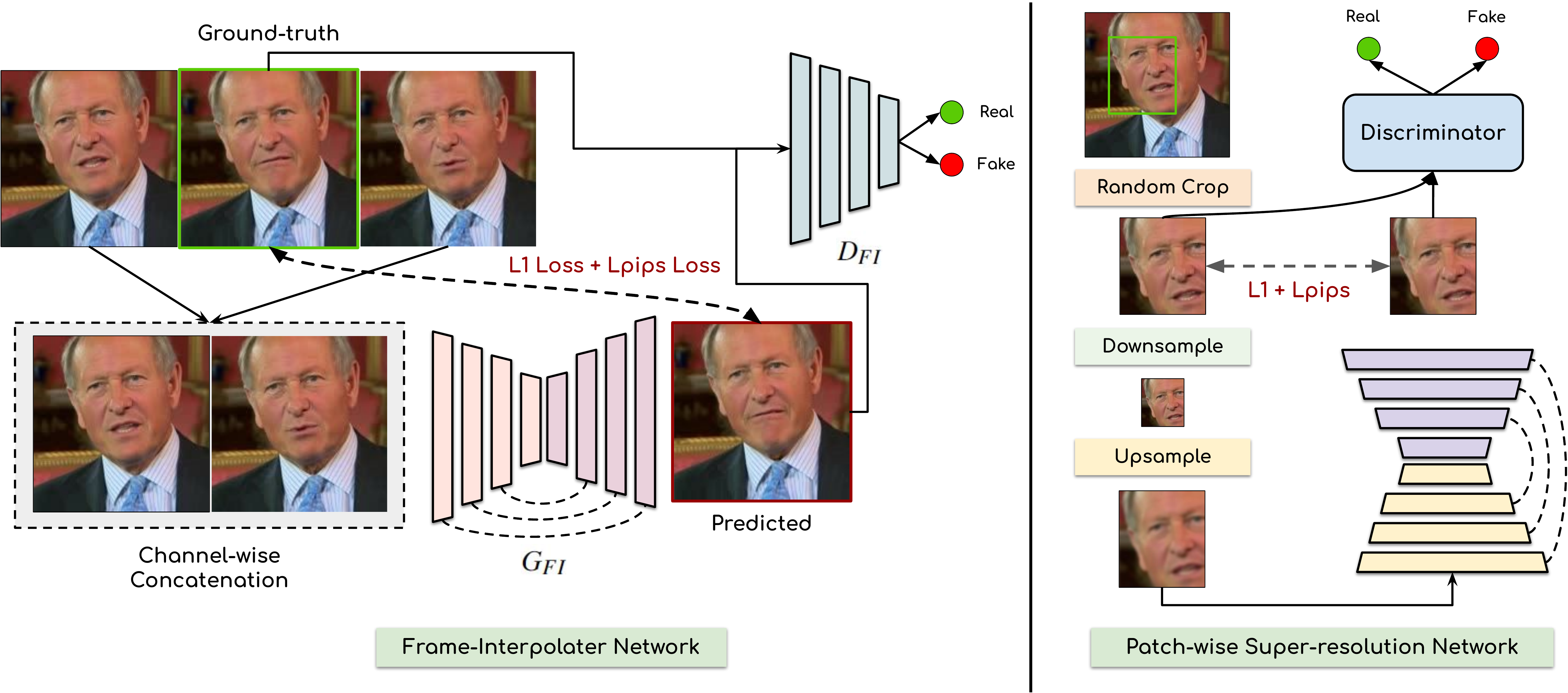}
    \caption{We depict the architectures of the frame-interpolation network and the Patch-wise Super-resolution Network.}
    \label{fig:interpolate_sr}
\end{figure*}

\vspace{-20pt}

\paragraph{Frame Interpolation at the Receiver's End}
To further reduce the bandwidth requirements and improve the compression ratio, we introduce a frame interpolation network motivated by recent advances in Face Enhancement works~\cite{gpen}. We use a standard GAN~\cite{goodfellow2014generative} architecture consisting of a Generator $G_{FI}$ and a Discriminator $D_{FI}$. To ensure lesser model complexity, we decide against using 3D convolution layers and use standard 2D convolution in both networks. As shown in Figure~\ref{fig:interpolate_sr}, we train this network on videos in a self-supervised manner. During training, we sample consecutive windows of three frames,  $\{v_i, v_{i+1}, v_{i+2}\}$ in a video. We then concatenate $v_i$ and $v_{i+2}$ channel wise creating the input to $G_{FI}$. The generator is tasked to generate $v_{i+1}$, which is used as the ground truth. The discriminator $D_{FI}$ is trained to maximize the loss function given in Equation~\ref{eq:disc_loss_interpolate}. We calculate three losses for the generator: the L1 reconstruction loss, the LPIPS~\cite{lpips} perceptual loss and finally, the discriminator's loss to train the generator $G_{FI}$. The loss and optimization functions used to train the generator are defined in Equation~\ref{eq:gen_loss_interpolate}. 

\begin{equation}
\begin{gathered}
v^{gen}_{i+1} = G_{FI}(v_i||v_{i+2})
\end{gathered}    
\end{equation}

\begin{equation}
\label{eq:disc_loss_interpolate}
\begin{gathered}
\max_{D_{FI}} L_{disc}(D_{FI}, G_{FI})=
\mathbb{E}_{real}[\log D_{FI}(v_{i+1})]
+ \mathbb{E}_{fake}[\log(1 - D_{FI}(v^{gen}_{i+1}))]
\end{gathered}    
\end{equation}

\begin{equation}
\label{eq:gen_loss_interpolate}
\begin{gathered}
\min_{G_{FI}} L_{gen} = L_{disc} + ||v_{i+1} - v^{gen}_{i+1}||_1  + LPIPS(v_{i+1}, v^{gen}_{i+1})
\end{gathered}    
\end{equation}


\paragraph{Patch-wise Super-resolution Network}
While users in the past were used to grainy webcam videos, the quality of the front cameras of cell phones, webcams, and other types of cameras has improved significantly. Maintaining the quality of the video calls is thus of utmost importance! Therefore, we train a GAN to enhance the quality and resolution of the generations. The architecture of this network closely resembles the frame interpolation network and is trained in a self-supervised manner. We also want our network to be able to arbitrarily super-resolve the output to any resolution. Therefore, instead of training the network on a fixed resolution of images, we train it using $k \times k$ cropped patches from the images. During training, we randomly sample frames from videos and take $k \times k$ random crops from them. We then bicubically downsample the patches by a random factor between $2 - 6$. We then create the input to the network by bicubically upsampling the downsampled patches back to their original resolution, i.e., $k \times k$. The network is tasked to remove the blur in the input patches introduced by bicubic upsampling. This network is also trained following a similar strategy to Equations~\ref{eq:disc_loss_interpolate} and~\ref{eq:gen_loss_interpolate}. During inference, we bicubically upsample the whole image to any desired resolution. Using a sliding window, we then divide the image into $k \times k$ patches. Our network then super-resolves each patch separately to generate sharp, high-resolution outputs. A pictorial representation of the architecture is given in Figure~\ref{fig:interpolate_sr}.


\paragraph{Adaptive Pivot Frame Selection}
Due to the lack of Jacobians in our key points, our network sometimes falters when faced with massive changes in head pose. While this is unlikely to happen in a dataset, it can be pretty standard when tested in a real-world video calling setup. We, therefore, propose a simple algorithm to adaptively change the pivot frame based on the difference in head pose and change in background. To detect the change in the head pose, we use an open-source codebase\footnote{\url{https://github.com/WIKI2020/FacePose_pytorch/}} and calculate the yaw, roll, and pitch in the pivot frame $F_{pv}$ and any current frame $F_i$ whose key points are to be transmitted. We empirically find thresholds of $\gamma_{yaw}, \gamma_{roll}, \gamma_{pitch}$ based on which we change the pivot frame to the current frame, i.e., $F_{pv} = F_i$ in case of a major shift. We also use Mediapipe~\cite{mediapipe} library to generate face segmentation masks to detect the background portions of a frame. We then use a pre-trained VGG-19~\cite{vgg} network to generate embeddings for the backgrounds of both $F_{pv}$ and $F_i$. A simple euclidean distance $d_{bg}$ is calculated to determine the amount of background change. If a significant background change is determined using an empirical threshold, the pivot frame is replaced by the current frame.

\paragraph{Dataset \& Implementation Details}
We train our networks on the train set from the VoxCeleb dataset~\cite{voxceleb} with a learning rate of $0.001$ using the Adam optimizer~\cite{adam}. The resolution of all the videos is kept at $256 \times 256$ during training. The patch size used for training the Super-Resolution network is set to $64 \times 64$. During inference, we apply $2\times$ super-resolution achieving $512 \times 512$ resolution on the final generated videos. The thresholds that we select after experimentation are $\gamma_{yaw} > 15^{\circ}, \gamma_{roll} > 15^{\circ}, \gamma_{pitch} > 15^{\circ}$. We select $d_{bg} > 0.05$ as the threshold for the considering backgrounds as different. Please note that breaching either of the thresholds is considered a criterion for replacing the pivot frame.  

\section{Experiments and Results}

\begin{wrapfigure}{r}{0.6\linewidth}
\centering
\includegraphics[width=1\linewidth]{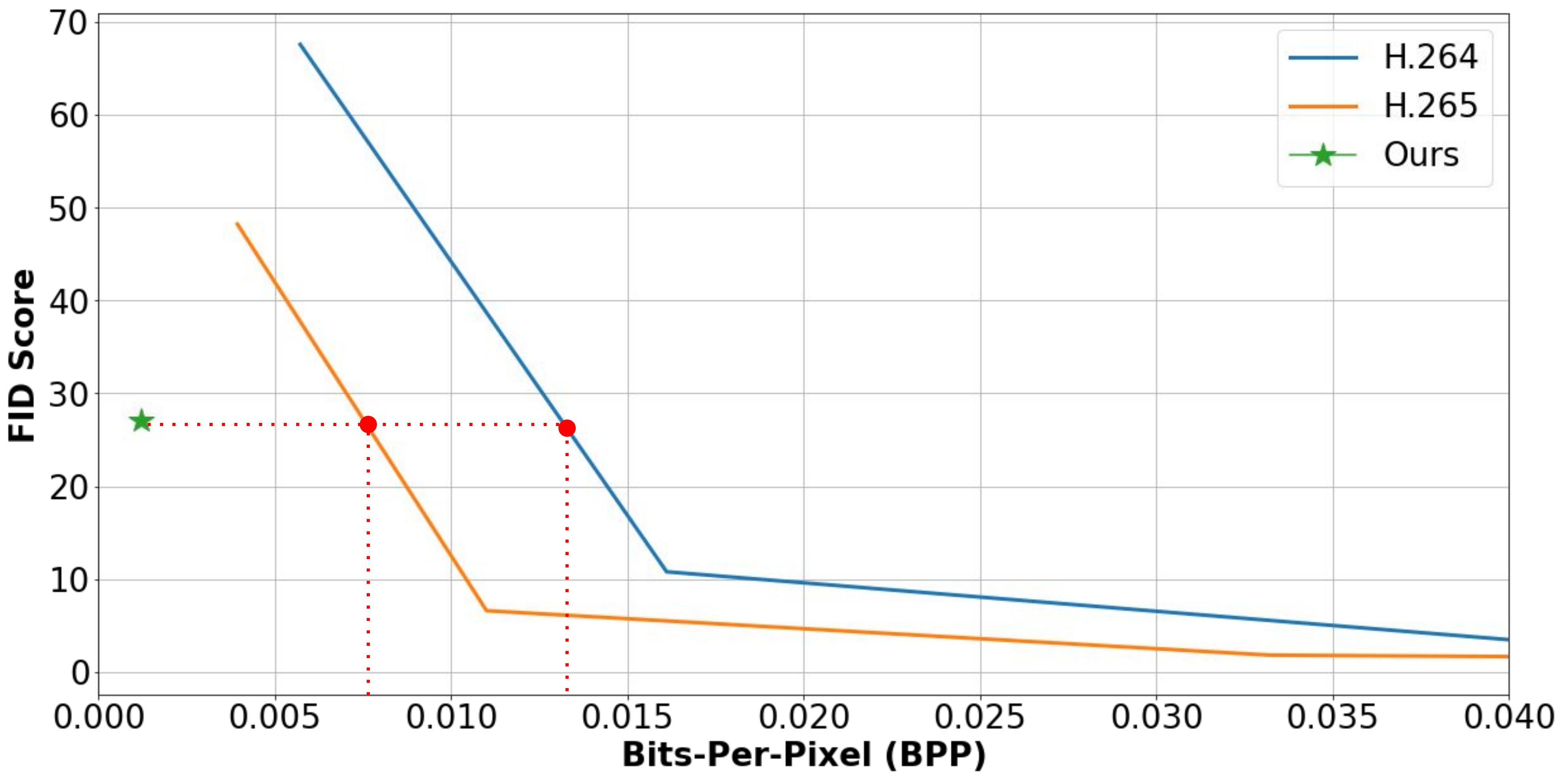}
\caption{We calculate the change of FID with reducing compression, i.e., increasing BPP. We find that the FID score achieved by our network can only be achieved at a far lower level of compression for both H.264 and H.265.}
\label{fig:compression_graph}
\end{wrapfigure}

\paragraph{Comparable Methods \& Metrics used}
We compare our work with two of the most famous and versatile video compression techniques, H.264~\cite{h_264} and H.265~\cite{h_265}. We vary the Constant Rate Factor (CRF) in both methods and generate results in various settings. We also compare our method with the original FOMM~\cite{Siarohin_2019_NeurIPS} and Face-Vid2Vid~\cite{wang2021one}. All the networks were trained on the same training set for a fair comparison. Apart from other comparable works, we also create baselines by removing different modules from our proposed pipeline. We report three visual quality metrics to measure the visual quality; PSNR, SSIM, and FID~\cite{fid}. We also report the BPP to measure the compression level for each method. We use the test set from the VoxCeleb~\cite{voxceleb} for benchmarking all the approaches. Please note that the BPP is calculated based on $512\times512$ as the final resolution for all the methods. 

\begin{figure*}[h]
    \centering
    \includegraphics[width=0.9\textwidth]{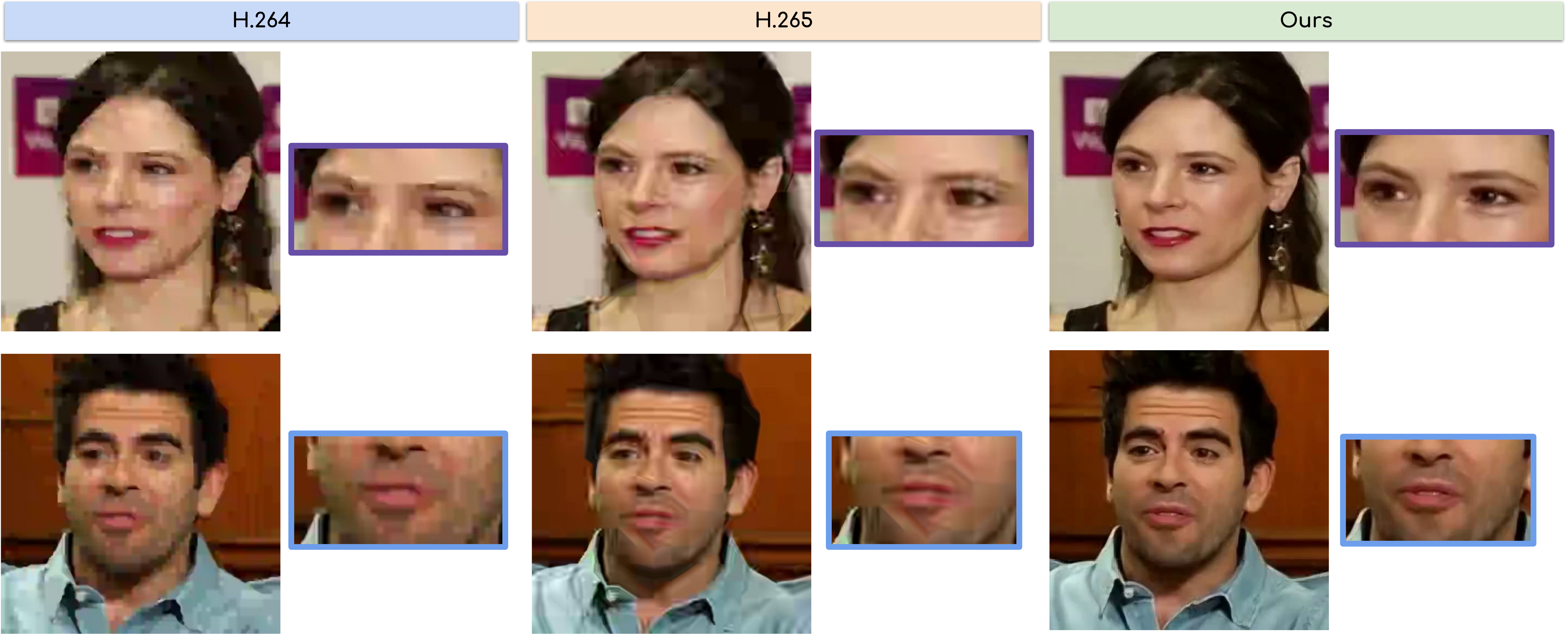}
    \caption{We compare our results with H.264 and H.265. Our method generates far sharper images with much less data.}
    \label{fig:h264_h265_compare}
\end{figure*}

\paragraph{Quantitative Results}

We report quantitative scores in Table~\ref{tab:fomm_compression}. For H.264 and H.265, we use the Constant Rate Factor (CRF) $= 51$ to generate the results at the least BPP possible. As we can see, even at the maximum compression levels of H.264 and H.265, our approach achieves less than $\frac{1}{3}$rd the BPP while generating visually appealing results. We also plot the variation of FID with changing BPP for H.264 and H.265 in Figure~\ref{fig:compression_graph}. We find that the FID achieved by our approach is only achievable at $5\times$ more BPP for H.265 and $10\times$ more BPP for H.264. We also achieve a much lower BPP than the original FOMM and Face-Vid2Vid~\cite{wang2021one} but can maintain quality. Finally, we stack up different modules from our approach one by one and then compare the results achieved in each combination. As observed in Table~\ref{tab:fomm_compression}, adding each module in our approach reduced BPP while maintaining the quantitative metrics at a similar level.





\begin{table*}[h]
\footnotesize
\begin{center}

\begin{tabular}{|c| c |c |c |c|} 
 \hline
 \textbf{Method} & \textbf{BPP$\downarrow$} & \textbf{PSNR$\uparrow$} & \textbf{SSIM$\uparrow$} & \textbf{FID$\downarrow$}\\ 
 \hline\hline
 FOMM  \cite{Siarohin_2019_NeurIPS} & 0.029 & 26.81 & 0.79 & 26.81 \\ 
 Face-vid2vid \cite{wang2021one} &  0.016  & 24.37 & 0.80 & 25.19\\ 
H.264  \cite{h_264} & 0.0057 & 30.25 & 0.78 &	67.54 \\
H.265  \cite{h_265} &  0.0039 & \textbf{30.74}	& 0.80	& 48.20 \\
\textbf{Key point Only (Ours)} & 0.0097 & 24.48 & 0.78 &	\textbf{20.58}\\

\textbf{Key point + Frame Interpolation (Ours)} & 0.0048 & 24.21 &	0.78 &	23.03\\

 \textbf{Key point + Frame Interpolation + SR (Ours)} & \textbf{0.0012} & 26.73 & \textbf{0.81} & 27.81 \\  
 \hline

\end{tabular}
 \caption{We compare our method with other state-of-the-art architectures as well as widely used techniques like H.264 and H.265. We observe our method to consistently have decent visual quality at much lower BPP.}
\label{tab:fomm_compression}
\end{center}
\end{table*}

\vspace{-30pt}

\begin{figure*}[h]
    \centering
    \includegraphics[width=\textwidth]{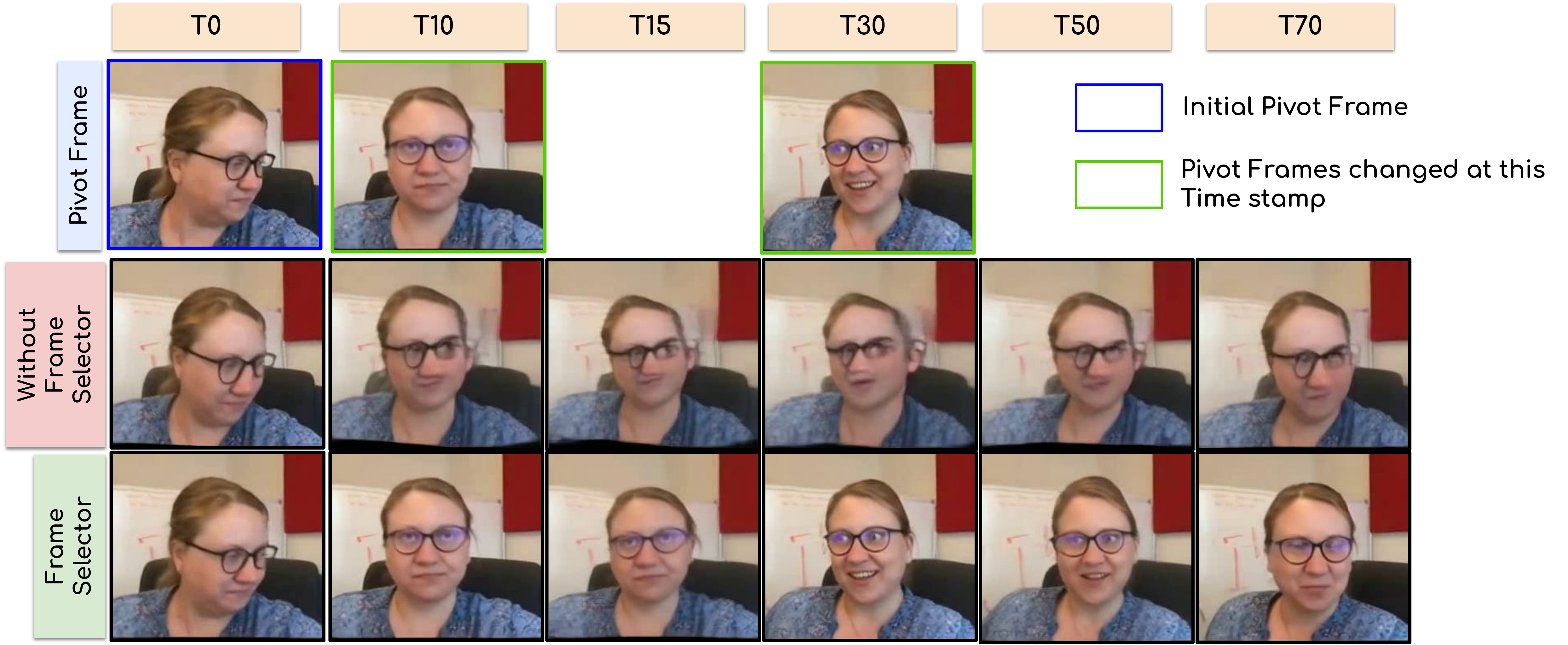}
    \caption{We use a lengthy real-world video call and mark frames for various time stamps (frame numbers in this case). Our goal is to understand the effect of the adaptive frame selector. In the above example, we select newer pivot frames at T10 and T30 owing to major head pose changes. This allows our network to continue generating sharp results.}
    \label{fig:real_usecase}
\end{figure*}

\paragraph{Qualitative Results}
We show multiple qualitative comparisons in Figure~\ref{fig:h264_h265_compare}. As we can see, our method generates sharper results when compared to the prevalent compression techniques. Furthermore, we also ran our pipeline on real video calls publicly available on YouTube. These videos are far longer than the ones present in the test set. Figure~\ref{fig:real_usecase} shows the impact of the adaptive pivot frame selection module and generates better outputs than the ones generated without using it.

\section{Ablation Studies}
\label{sec:ablation}

We perform several ablation studies to understand the effectiveness of different hyperparameters we choose to achieve the best performance. We keep the pivot frame constant for all the experiments if not mentioned otherwise. 

While we train our network to interpolate a single frame at a time, it can be easily used to interpolate more than one frame during inference by using the generated frames as input. We interpolate $2$ frames and $3$ frames at a time and report the scores in Table~\ref{tab:interpolate}. As we see, interpolating more number of frames improves the BPP significantly but also leads to some loss in performance. However, the performance still remains within usable range and thus can be explored in cases where even more compression is required.

\begin{table*}[h]
\footnotesize
\begin{center}
\begin{tabular}{|c| c |c |c |c|} 
 \hline
 \textbf{\#Int. frames} & \textbf{BPP$\downarrow$} & \textbf{PSNR$\uparrow$} & \textbf{SSIM$\uparrow$} & \textbf{FID$\downarrow$}\\ 
 \hline\hline
 0  & 0.0024 & 29.28	& 0.83	& 9.10 \\
 1  & 0.0012 & 28.49	& 0.82	& 12.42 \\
 2  & 0.0008 & 28.23 & 0.81	& 12.77 \\
 3  & 0.0006 & 27.73 & 0.79	& 12.92 \\
 \hline

\end{tabular}
 \caption{We vary the number of frames that are interpolated and report the scores achieved.}
  \label{tab:interpolate}

\end{center}
\end{table*}

We also vary the $k \times k$ patch size used for training the super resolution network. We super resolve the output $2$ times using different sized patches and report our findings in Table~\ref{tab:sr_patch}.

\begin{table*}[h]
\footnotesize
\begin{center}

\begin{tabular}{|c|c |c |c|} 
 \hline
 \textbf{Patch Size}& \textbf{PSNR$\uparrow$} & \textbf{SSIM$\uparrow$} & \textbf{FID$\downarrow$}\\ 
 \hline\hline
 $128\times128$ &  27.37	& 0.81 & 19.12 \\
 $64\times64$ & 27.47	& 0.80 & 20.16 \\
 $32\times32$ & 26.18 & 0.79 & 20.89\\
 $16\times16$ & 25.34 & 0.78 & 21.17\\
 \hline

\end{tabular}

 \caption{We vary the size of the patches taken by our SR network and report the scores in this table.}
  \label{tab:sr_patch}
\end{center}
\end{table*}
We select different thresholding parameters for our frame selection algorithm. We report the scores in Table~\ref{tab:adaptive}. Using adaptive frame selection increases BPP due to the transfer of multiple pivot frames. We calculate the metrics using different thresholds for all the $\gamma$ variables and $d_{bg}$. On an average, at our default setting of $\gamma>15^{\circ}$ and $d_{bg} > 0.05$, we find a pivot frame change every $10$ seconds. The shift is much less common on the bigger thresholds.    

\begin{table*}[h]
\footnotesize
\begin{center}
\begin{tabular}{|c| c |c |c |c|} 
 \hline
 \textbf{Method} & \textbf{BPP$\downarrow$} & \textbf{PSNR$\uparrow$} & \textbf{SSIM$\uparrow$} & \textbf{FID$\downarrow$}\\ 
 \hline\hline
$d_{bg} > 0.05$ & 0.0029 & 27.37 & 0.81 & 19.12 \\
$d_{bg} > 0.06$ & 0.0021 & 25.93& 0.77 & 20.74 \\
$d_{bg} > 0.07$ & 0.0016 & 24.72 & 0.73 & 23.17 \\
$\gamma > 15^{\circ} $ & 0.0049 & 25.28 & 0.75 & 22.46 \\
$\gamma > 30^{\circ}$ & 0.0031 & 24.02	& 0.71 & 26.63 \\
$\gamma > 45^{\circ}$ & 0.0018 & 23.76	& 0.70 & 26.93 \\

\hline
\end{tabular}

 \caption{We select different thresholds for our adoptive frame selection algorithm. Please note that $\gamma$ here represents thresholds for all the three $\gamma$-values.}
 \label{tab:adaptive}
\end{center}
\end{table*}

\vspace{-35pt}
\section{Conclusion}
In this work, we propose to use the high-level semantics of a talking head video to create extreme compression schemes which can revolutionize video calling. Our work uses compact key points to transmit information about the talking head in each video frame. We also propose a frame interpolation network followed by super-resolution to arbitrary resolutions. Finally, a pivot frame selection algorithm is used for long video calls helping our compression technique continue generating high-quality videos. In the future, we believe solving other aspects like ensuring its application on edge devices will be a prospective task.

\textbf{Acknowledgement:} This work is partly supported by Huawei, India



\bibliography{egbib}
\end{document}